\documentclass[11pt]{article}

\usepackage[final]{acl}

\usepackage{times}
\usepackage{latexsym}
\usepackage[T1]{fontenc}
\usepackage[utf8]{inputenc}
\usepackage{microtype}
\usepackage{inconsolata}
\usepackage{placeins}
\usepackage{amsmath}
\usepackage{amssymb}
\usepackage{subcaption}

\usepackage{graphicx}
\usepackage{multirow}
\usepackage{booktabs}
\usepackage{siunitx}
\usepackage{diagbox}
\usepackage[table]{xcolor}
\usepackage{colortbl}
\usepackage{float}
\usepackage{titletoc}

\title{REVEAL: Reference-Grounded Reasoning for Multimodal \\ Manipulation Detection}

\author{
Jun Zhou \quad Bingwen Hu\thanks{\ \ Corresponding author.} \quad Yaxiong Wang \quad Zhedong Zheng \quad
Yongzhen Wang \\[0.3cm] \textbf{Yuchen Zhang \quad Ping Liu}
}

\begin{document}
\maketitle

\begin{abstract}
Multimodal manipulation detection aims to simultaneously identify forged image--text pairs and localize tampered regions, yet existing methods typically rely on memorizing isolated artifacts and struggle with imperceptible manipulation traces or domain shifts.
Inspired by human comparative reasoning, we reformulate this task as a \emph{reference-grounded verification} problem, where authenticity is assessed by comparing a query against retrieved authentic evidence.
We propose \textbf{REVEAL} (\textbf{R}eference-\textbf{E}nabled \textbf{V}erification for \textbf{E}vidence \textbf{A}nalysis and \textbf{L}ocalization), a framework explicitly designed for this comparative paradigm.
To support this paradigm, we construct a large-scale reference library comprising 170K authentic news image--text pairs featuring over 40K public figures.
Technically, REVEAL employs a difference-aware fusion mechanism to capture fine-grained discrepancies between the query and retrieved evidence. 
Furthermore, we introduce a task-decoupled Mixture-of-Experts (MoE) architecture to jointly execute instance-level detection and fine-grained grounding, effectively mitigating optimization conflicts between these heterogeneous objectives.
Extensive experiments demonstrate that REVEAL significantly outperforms state-of-the-art methods, and notably enables \emph{training-free domain adaptation} by simply updating the reference library, offering a robust and practical solution for detecting evolving misinformation. Code is available at \textcolor{magenta}{\href{https://anonymous.4open.science/r/REVEAL-Reference-A006}{\nolinkurl{https://anonymous.4open.science/r/REVEAL-Reference-A006}}.}

\end{abstract}

\begin{figure}[t] 
  \centering

  \includegraphics[width=\columnwidth]{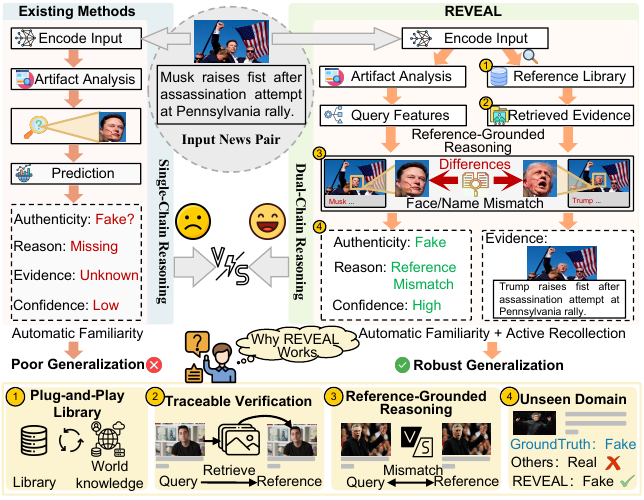} 
\caption{
\textbf{Artifact-centric vs. Reference-grounded reasoning}. While existing methods rely on isolated, artifact-centric cues that often yield opaque and low-confidence predictions, REVEAL shifts to a reference-grounded paradigm. By mimicking active recollection via a plug-and-play memory, REVEAL explicitly contrasts the input against retrieved authentic evidence. This mechanism inherently exposes semantic inconsistencies (e.g., face/name mismatches), delivering highly interpretable decisions with traceable manipulation reasoning and robust generalization to unseen domains.}
  \label{fig:introduction}
\end{figure}

\section{Introduction}
The rapid progress of generative models has made fabricated media increasingly realistic, posing serious threats to information authenticity and public trust~\citep{huang-etal-2024-miragenews,yan-etal-2025-trust,zhang-etal-2025-generation}.
Among different misinformation formats, image--text news pairs are especially important due to their wide use on social platforms~\citep{zeng-etal-2024-multimodal,rahman-etal-2025-exploring,tonglet-etal-2024-image}, where effective analysis demands multi-granularity reasoning to determine authenticity, identify manipulation types, and localize tampered regions across modalities~\citep{hammer,hammer++}.

However, two limitations are emerging for existing detectors. First, visual artifacts that once served as reliable cues, such as boundary inconsistencies or frequency anomalies, are becoming increasingly imperceptible as generative models improve~\citep{corvi2023detection,ricker2024detection}. Second, detectors that memorize training-specific artifact patterns generalize poorly to unseen news sources or novel manipulation pipelines~\citep{cozzolino2024raising,yan2024transcending}.
This raises a key question: when intrinsic artifacts are neither perceptible nor transferable, what other signal can be exploited?
Inspired by human studies showing that deepfake judgments are influenced not only by cognitive tendencies such as truth bias, but also by prior knowledge when familiar figures or events are involved~\citep{Groh,diel2024human,somoray2025human}, we introduce a comparison-based verification paradigm as a complementary reasoning pathway.
For example, a manipulated news photograph showing Elon Musk raising his fist after an assassination attempt would be questioned not because of visual artifacts, but because the authentic event involved Donald Trump.
To operationalize this verification pattern, Figure~\ref{fig:introduction} distinguishes between two reasoning modes: \emph{Automatic Familiarity}, which judges plausibility through internal artifact analysis, and \emph{Active Recollection}, which validates the query against external authentic evidence. Most existing methods rely solely on the former~\citep{cozzolino2024clip,yan2024transcending,cui2025forensics}. By performing single-chain reasoning purely from learned artifact patterns, they often yield opaque decisions and exhibit brittle behavior under distribution shifts.

In contrast, REVEAL augments automatic familiarity with active recollection via an information retrieval paradigm. Here, an external memory serves as an authentic reference library, where recollection is driven by similarity-based retrieval, and verification proceeds through cross-reference reasoning. Given a query, our framework synthesizes internal artifact cues with retrieved references. By applying difference-aware reasoning across these signals, it effectively exposes visual, textual, and cross-modal inconsistencies. Consequently, REVEAL delivers authenticity predictions grounded in interpretable evidence, explicitly highlighting the retrieved references and detected mismatches.
Crucially, this dual-chain architecture inherently resolves the generalization bottleneck. Because the reference library updates independently of the model parameters, REVEAL achieves training-free domain adaptation, seamlessly handling unseen news sources or novel manipulation pipelines.

To realize the above reference-grounded verification paradigm, two challenges must be addressed: constructing an informative authentic reference library and effectively integrating retrieved evidence for multi-granularity detection. 
We therefore present \textbf{REVEAL} (\textbf{R}eference-\textbf{E}nabled \textbf{V}erification for \textbf{E}vidence \textbf{A}nalysis and \textbf{L}ocalization), which retrieves authentic image--text evidence at inference time and verifies the query through cross-reference comparison. 
Built upon a reference library of 170K authentic news image--text pairs covering over 40K public figures, REVEAL employs difference-aware fusion and a task-decoupling Mixture-of-Experts architecture~\citep{chen2023adamv,shazeer2017outrageously,lepikhingshard} to support both global detection and fine-grained grounding.

In summary, our main contributions are as follows:

\begin{itemize}
    \item We propose Reference-Grounded Verification, shifting the paradigm of multimodal forensics from isolated artifact memorization to retrieval-conditioned comparative reasoning. 
    
    \item We construct a comprehensive reference library comprising 170K authentic news image-text pairs across over 40K public figures. This foundation preserves high-fidelity cross-modal correspondences, enabling scalable, evidence-driven verification.
    
    \item We develop a difference-aware fusion mechanism to capture fine-grained query-evidence discrepancies, alongside a task-decoupled Mixture-of-Experts (MoE) architecture that jointly handles global detection and fine-grained grounding.
    
    \item Extensive evaluations demonstrate that REVEAL consistently outperforms existing methods in both detection and localization. Crucially, our paradigm achieves training-free domain adaptation simply by updating the external reference library.
\end{itemize}

\begin{figure*}[t]
  \centering
  \includegraphics[width=\textwidth]{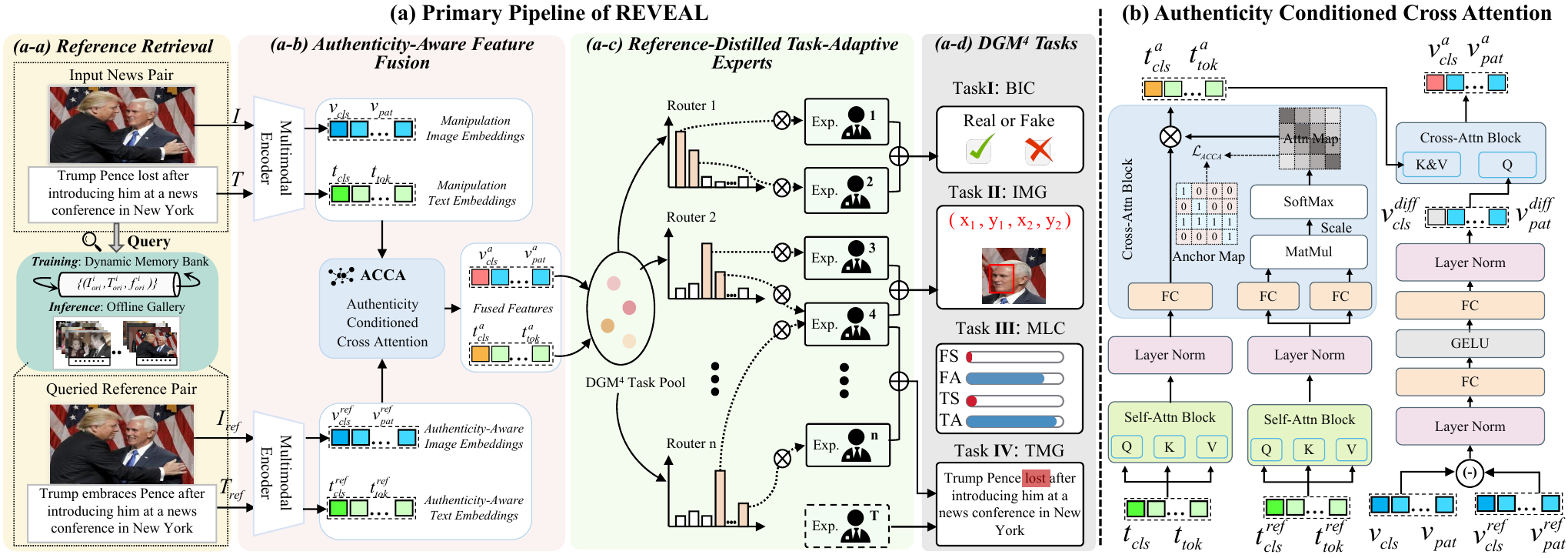}
\caption{\textbf{Overview of the proposed REVEAL framework.}
\textbf{(a) Primary Pipeline:}
(a-a) \textit{Reference Retrieval obtains semantically related reference pairs from a dynamic memory bank during training or an offline retrieval gallery during inference};
(a-b) \textit{Authenticity-Aware Feature Fusion} employs the proposed ACCA module to fuse query and reference features, producing authenticity-conditioned representations;
(a-c) \textit{Reference-Distilled Task-Adaptive Experts} leverages a router-based Mixture-of-Experts architecture within the DGM$^4$ task pool, where each expert specializes in different manipulation patterns;
(a-d) \textit{DGM$^4$ Tasks} include four objectives: binary classification (BIC), image manipulation grounding (IMG), multi-label classification (MLC), and text manipulation grounding (TMG).
(b) Detailed architecture of Authenticity Conditioned Cross Attention (ACCA),
which computes cross-modal attention between query and reference to produce authenticity-aware representations.}
\label{fig:framework}
\end{figure*}

\section{Related Work}
\subsection{Retrieval-Augmented Learning}

Retrieval-augmented learning incorporates external evidence at inference time to reduce reliance on parametric knowledge.
In NLP, methods improve factual grounding via self-reflective or corrective retrieval, e.g., Self-RAG~\citep{asai2024selfrag} and CRAG~\citep{yan2024crag}, while others refine when and how retrieved evidence is used, such as RankRAG~\citep{yu2024rankrag} and FLARE~\citep{jiang2023flare}.
Multimodal extensions further ground visual reasoning with retrieved knowledge, e.g., MuRAG~\citep{chen2022murag} for image--document reasoning and Wiki-LLaVA~\citep{caffagni2024wikilava} for knowledge-grounded VQA, with recent work exploring retrieval over visual documents~\citep{yu2024visrag} and knowledge-guided multimodal generation~\citep{yang2024rag}.
However, these approaches use retrieval to \emph{enrich generation}.
In contrast, retrieval in our framework serves \emph{verification}: retrieved references act as authentic anchors for exposing contradictions, inconsistencies, and manipulations, rather than augmenting output content.

\subsection{Multimodal Manipulation Detection}
Early manipulation detection focused on unimodal forgery, particularly face manipulation, leveraging spatial artifacts, frequency inconsistencies, and temporal coherence to identify deepfakes~\citep{rossler2019faceforensics,li2020face}.
As real-world misinformation increasingly involves coordinated cross-modal manipulation~\citep{hu2023read}, recent work has shifted toward multimodal detection.
HAMMER~\citep{hammer} performs joint detection and grounding via shallow cross-modal fusion with modality-specific decoders.
Building on this, ASAP~\citep{asap} introduces hierarchical cross-modal alignment to capture subtle manipulation traces, while UFAFormer~\citep{ufa} incorporates frequency-aware analysis to expose generation artifacts imperceptible in the spatial domain.
More recently, IDseq~\citep{idseq} adopts a decoupled sequential pipeline that first detects manipulation at the instance level and then grounds manipulated regions in each modality separately.
Despite these advances, existing methods fundamentally operate in an \emph{instance-isolated} manner: each sample is judged primarily from patterns learned during training, without access to authentic external references for comparative verification~\citep{cozzolino2024raising,yan2024transcending}.

\section{Methodology}

\subsection{Problem Formulation}

We formulate multimodal manipulation detection as reference-grounded verification.
Let $(I,T)$ denote a query image--text pair and 
$\mathcal{R}=\{(I_k,T_k)\}_{k=1}^{N}$ an authentic reference library.
Instead of judging $(I,T)$ in isolation, our detector conditions its prediction on retrieved evidence:
\begin{equation}
\mathcal{D}_{\theta}(I,T;\mathcal{R})
=
\mathcal{D}_{\theta}(I,T \mid I_{\mathrm{ref}},T_{\mathrm{ref}}),
\label{eq:eq1}
\end{equation}
where $(I_{\mathrm{ref}},T_{\mathrm{ref}})\in\mathcal{R}$ is the retrieved authentic reference.
Thus, manipulation cues are modeled as query-reference inconsistencies rather than absolute artifacts within a single sample.

Figure~\ref{fig:framework} shows the overall architecture of \textbf{REVEAL}.
Given a query image--text pair $(I,T)$, REVEAL first retrieves semantically relevant authentic references $(I_{\mathrm{ref}},T_{\mathrm{ref}})$ from a reference library, then compares the query with the retrieved evidence through an authenticity-aware fusion module, and finally feeds the fused discrepancy features into a Mixture-of-Experts predictor for authenticity verification and multimodal grounding.
Through this pipeline, REVEAL detects manipulations by reasoning over query-reference inconsistencies rather than isolated artifact patterns.

\subsection{Reference Retrieval}

As shown in Figure~\ref{fig:framework}(a-a), REVEAL retrieves semantically relevant authentic references from a dynamic memory bank during training and an offline gallery at inference. 
The reference gallery is plug-and-play: for domain-specific detection, it can be constructed from target news sources, while for general-purpose detection, we build a universal gallery containing 170K authentic image--text pairs covering over 40K public figures.
This design enables training-free domain adaptation by updating the gallery without modifying model parameters.

During training, we maintain a memory bank of authentic pairs with cached visual embeddings. Crucially, to guarantee strict data isolation and prevent any potential data leakage, each sample in the training set is exclusively paired with its corresponding authentic reference from the training partition.

For each query image $I_q$, candidate references are retrieved by top-$K$ cosine similarity:
\begin{equation}
\mathcal{K}
=
\operatorname{TopK}_{j}
\cos(\phi(I_q), \mathbf{f}_j),
\label{eq:retrieval}
\end{equation}
where $\phi(\cdot)$ is the visual encoder and $\mathbf{f}_j$ denotes the cached embedding of the $j$-th reference.
To improve retrieval quality, we impose a Visual Retrieval Contrastive loss $\mathcal{L}_{\text{VRC}}$ following InfoNCE~\citep{oord2018representation}, which encourages queries to be close to their authentic counterparts in the embedding space.

At inference time, the selected reference library is indexed as an offline retrieval gallery, and references are retrieved using FAISS~\citep{johnson2019billion} for efficient approximate nearest-neighbor search over large-scale authentic evidence.

\subsection{Authenticity-Aware Feature Fusion}

The retrieved reference provides an authentic anchor for detecting query-reference discrepancies.
As shown in Figure~\ref{fig:framework} (a-b), we propose Authenticity Conditioned Cross Attention (ACCA) to compare query and reference features and produce discrepancy-aware representations.

For textual comparison, query caption tokens attend to the retrieved reference caption, yielding a discrepancy-aware textual representation:
\begin{equation}
\mathbf{T}_{\mathrm{diff}}
=
\operatorname{CrossAttn}
(\mathbf{T}_q,\mathbf{T}_{\mathrm{ref}},\mathbf{T}_{\mathrm{ref}}),
\end{equation}
where $\mathbf{T}_{\mathrm{diff}} = [t^a_{\mathrm{cls}}, t^a_{\mathrm{tok}}]$, $\mathbf{T}_q = [t_{\mathrm{cls}}, t_{\mathrm{tok}}]$ and $\mathbf{T}_{\mathrm{ref}} = [t^{\mathrm{ref}}_{\mathrm{cls}}, t^{\mathrm{ref}}_{\mathrm{tok}}]$.
We further introduce Reference-Guided Attention Supervision (RGAS) to regularize this attention, encouraging authentic tokens to align with matched reference evidence while guiding manipulated tokens to search broader reference context. The RGAS loss is formulated as:
\begin{equation}
\label{eq:rgas}
    \mathcal{L}_{\text{RGAS}} = \frac{1}{|\mathcal{V}|} \sum_{(i,j) \in \mathcal{V}} \ell(A_{i,j}, G_{i,j}),
\end{equation}
where $\mathcal{V}$ denotes valid non-padding positions.

For the visual modality, we retrieve the top-ranked authentic sample and use it as the reference anchor for discrepancy modeling. We model anchor-relative discrepancies by comparing the query image with the retrieved reference in the feature space. 
Specifically, the query and reference features are compared through element-wise subtraction and transformed into manipulation-aware visual representations.
Finally, textual and visual discrepancy features are fused through cross-modal attention, enabling the model to jointly reason over semantic and visual evidence for detection and grounding.

\subsection{Reference-Distilled Task-Adaptive Experts}
\label{sec:method}
As shown in Figure~\ref{fig:framework}(a-c), we design a Reference-Distilled Task-Adaptive Experts architecture to address the multi-granularity demands of multimodal verification. Detection tasks rely on holistic semantic reasoning, while grounding tasks require fine-grained local alignment. We therefore adopt an MoE architecture with task-aware routing, where explicit task identifiers guide experts to specialize in task-specific feature patterns.

\noindent\textbf{Expert Network Design.}
To address the heterogeneous demands of reference-grounded verification, we define a task pool $\mathcal{T}=\{\mathrm{BIC}, \mathrm{IMG}, \mathrm{MLC}, \mathrm{TMG}\}$, as illustrated in Figure~\ref{fig:framework}. Recognizing that classification and grounding inherently require distinct feature granularities, we allocate a dedicated pool of $N$ experts to each task $t \in \mathcal{T}$. This isolation enables classification-oriented experts to extract holistic representations, while grounding-oriented experts isolate fine-grained spatial or token-level cues. Formally, each expert is instantiated as an independent feed-forward network:
\begin{equation}
\mathcal{E}_n^{(t)}(\mathbf{x}) = \mathrm{FFN}_n^{(t)}(\mathbf{x}), \quad n \in \{1,\ldots,N\}.
\end{equation}

This task-decoupled design allows each expert pool to learn task-specific transformations and mitigates interference between global detection and fine-grained grounding objectives.

\noindent\textbf{Task-Aware Routing.}
To address heterogeneous sub-task requirements, we adopt a task-aware MoE head whose routing is conditioned on both input features and task identity. For each task $t \in \mathcal{T}$, we use an independent expert pool $\{\mathcal{E}^{(t)}_n\}_{n=1}^{N}$ and a learnable task embedding $\mathbf{e}_t \in \mathbb{R}^{d/4}$. The MoE head is applied separately to each task with input $\mathbf{X}^{(t)} \in \mathbb{R}^{L_t \times d}$, where $L_t=1$ for BIC, IMG, and MLC using $\mathbf{x}_{\mathrm{cls}}$, and $L_t=L$ for TMG using $\mathbf{X}_{\mathrm{tok}}$.
The router mean-pools the input and concatenates it with the task embedding:
\begin{equation}
\mathbf{h}^{(t)} =
\operatorname{MeanPool}(\mathbf{X}^{(t)}) \oplus \mathbf{e}_t .
\end{equation}

A lightweight gating network then produces the routing distribution over the $N$ experts:
\begin{equation}
\mathbf{p}^{(t)} =
\operatorname{softmax}\left(\operatorname{Gate}(\mathbf{h}^{(t)})\right).
\end{equation}
We select the top-$K$ experts according to $\mathbf{p}^{(t)}$ and re-normalize their weights:
\begin{equation}
\tilde{p}^{(t)}_n =
\frac{p^{(t)}_n}
{\sum_{j \in \mathcal{S}^{(t)}} p^{(t)}_j},
\quad n \in \mathcal{S}^{(t)} ,
\end{equation}
where $\mathcal{S}^{(t)}=\operatorname{TopK}(\mathbf{p}^{(t)})$.
The task output is obtained by aggregating the selected experts:
\begin{equation}
\mathbf{Y}^{(t)} =                                                
\sum_{n \in \mathcal{S}^{(t)}}
\tilde{p}^{(t)}_n
\mathcal{E}^{(t)}_n(\mathbf{X}^{(t)}).
\end{equation}

For BIC, MLC, and IMG, the MoE operates on sentence-level features to produce global classification or box regression outputs, while for TMG, it is applied to token-level features to generate manipulation logits for each token.
Thus, the router performs task-conditioned expert selection, and each task-specific expert pool learns features suited to its own reasoning granularity.

\noindent\textbf{Reference-Enhanced Expert Distillation.}
Text Manipulation Grounding (TMG) is challenging because manipulated tokens are sparse and can be weakened during multi-task optimization.
To improve token-level sensitivity, we introduce a reference-enhanced teacher for TMG.
Unlike the student head, the teacher directly takes the query-reference text pair $[\mathbf{T}_q;\mathbf{T}_{\mathrm{ref}}]$ as input, preserving fine-grained discrepancy cues between the query and authentic evidence.
The teacher's predictions are then distilled to the MoE student through a temperature-scaled KL divergence loss:
\begin{equation}
\mathcal{L}_{\mathrm{distill}}
=
\tau^2
D_{\mathrm{KL}}
\left(
\sigma(\mathbf{z}_{\mathrm{teacher}}/\tau)
\|
\sigma(\mathbf{z}_{\mathrm{student}}/\tau)
\right).
\end{equation}

By providing soft supervision, this distillation mechanism forces the student to maintain acute sensitivity to subtle textual manipulations.

\begin{table*}[t]
\centering
\caption{Performance comparison with state-of-the-art methods on DGM$^4$ dataset~\citep{hammer}.
\textbf{Bold} indicates best performance, and \underline{underlined} indicates second-best. $\uparrow$/$\downarrow$ indicates higher/lower is better, and $\Delta$AVG denotes the average improvement compared to HAMMER.}
\label{tab:DGM}
\setlength{\tabcolsep}{2pt}
\renewcommand{\arraystretch}{1.1}
\small
\resizebox{0.98\textwidth}{!}{
\begin{tabular}{@{}l *{12}{S[table-format=2.2]} c@{}}
\toprule
\multirow{2}{*}{\textbf{Methods}} &
\multicolumn{3}{c}{\textbf{Binary Cls}} &
\multicolumn{3}{c}{\textbf{Multi-Label Cls}} &
\multicolumn{3}{c}{\textbf{Image Grounding}} &
\multicolumn{3}{c}{\textbf{Text Grounding}} &
\multirow{2}{*}{\textbf{$\Delta$AVG}} \\
\cmidrule(lr){2-4} \cmidrule(lr){5-7} \cmidrule(lr){8-10} \cmidrule(lr){11-13}
& {AUC $\uparrow$} & {EER $\downarrow$} & {ACC $\uparrow$}
& {mAP $\uparrow$} & {CF1 $\uparrow$} & {OF1 $\uparrow$}
& {IoU\textsubscript{m} $\uparrow$} & {IoU\textsubscript{50} $\uparrow$} & {IoU\textsubscript{75} $\uparrow$}
& {Precision $\uparrow$} & {Recall $\uparrow$} & {F1 $\uparrow$}
& \\
\midrule
CLIP~\citep{radford2021learning} & 83.22 & 24.61 & 76.40 & 66.00 & 59.52 & 62.31 & 49.51 & 50.03 & 38.79 & 58.12 & 22.11 & 32.03 & \multicolumn{1}{c}{--} \\
VILT~\citep{kim2021vilt} & 85.16 & 22.88 & 78.38 & 72.37 & 66.14 & 66.00 & 59.32 & 65.18 & 48.10 & 66.48 & 49.88 & 57.00 & \multicolumn{1}{c}{--} \\
HAMMER~\citep{hammer} & 93.19 & 14.10 & 86.39 & 86.22 & 79.37 & 80.37 & 76.45 & 83.75 & 76.06 & 75.01 & 68.02 & 71.35 & 0.00 \\
HAMMER++~\citep{hammer++} & 93.33 & 14.06 & 86.66 & 86.41 & 79.73 & 80.71 & 76.46 & 83.77 & 76.03 & 73.05 & 72.14 & 72.59 & +0.35 \\
ViKI~\citep{li2024towards} & 93.51 & 13.87 & 86.67 & 86.58 & 81.07 & 80.10 & 76.51 & 83.95 & 75.77 & 77.79 & 66.06 & 72.44 & +0.38 \\
UFAFormer~\citep{ufa} & 93.81 & 13.60 & 86.80 & 87.85 & 80.31 & 81.48 & 78.33 & 85.39 & 79.20 & 73.35 & 70.73 & 72.02 & +1.13 \\
ASAP~\citep{asap} & 94.38 & 12.73 & 87.71 & 88.53 & 81.72 & 82.89 & 77.35 & 84.75 & 76.54 & \textbf{79.38} & \underline{73.86} & \textbf{76.52} & +2.40 \\
IDseq~\citep{idseq} & \underline{94.55} & \underline{11.40} & \underline{88.94} & \underline{90.01} & \underline{83.00} & \underline{84.90} & \underline{83.33} & \underline{89.39} & \underline{86.19} & 75.96 & 71.23 & 73.52 & \underline{+3.96} \\
\midrule
\textbf{REVEAL (Ours)} & \textbf{97.82} & \textbf{3.75} & \textbf{93.18} & \textbf{91.37} & \textbf{85.13} & \textbf{86.68} & \textbf{85.51} & \textbf{93.53} & \textbf{86.52} & \underline{79.31} & \textbf{74.45} & \underline{75.34} & \textbf{+6.92}\\
\bottomrule
\end{tabular}%
}
\end{table*}

\begin{table*}[!ht]
\centering
\caption{Comparison with state-of-the-art methods on the SAMM dataset using the original SAMM image reference gallery. \textbf{Bold} indicates best and \underline{underlined} indicates second-best. $\uparrow$/$\downarrow$ indicates higher/lower is better, and $\Delta$AVG denotes the average improvement compared to HAMMER.}
\label{tab:comparison}
\setlength{\tabcolsep}{2pt}
\renewcommand{\arraystretch}{1.15}
\small
\resizebox{\textwidth}{!}{
\begin{tabular}{@{}l *{12}{S[table-format=2.2]} c@{}}
\toprule
\multirow{2}{*}{\textbf{Methods}} &
\multicolumn{3}{c}{\textbf{Binary Cls}} &
\multicolumn{3}{c}{\textbf{Multi-Label Cls}} &
\multicolumn{3}{c}{\textbf{Image Grounding}} &
\multicolumn{3}{c}{\textbf{Text Grounding}} &
\multirow{2}{*}{\textbf{$\Delta$AVG}} \\
\cmidrule(lr){2-4} \cmidrule(lr){5-7} \cmidrule(lr){8-10} \cmidrule(lr){11-13}
& {AUC $\uparrow$} & {EER $\downarrow$} & {ACC $\uparrow$}
& {mAP $\uparrow$} & {CF1 $\uparrow$} & {OF1 $\uparrow$}
& {IoU\textsubscript{m} $\uparrow$} & {IoU\textsubscript{50} $\uparrow$} & {IoU\textsubscript{75} $\uparrow$}
& {Precision $\uparrow$} & {Recall $\uparrow$} & {F1 $\uparrow$}
& \\
\midrule
VILT~\citep{kim2021vilt}                & 96.10 & 11.02 & 88.83 & 96.03 & 90.21 & 89.84 & 65.38 & 71.91 & 54.49 & 77.42 & 69.78 & 73.40 & \multicolumn{1}{c}{--} \\
HAMMER~\citep{hammer}                   & 97.85 &  7.80 & 92.43 & 97.98 & 93.77 & 93.44 & 77.68 & 84.41 & 78.44 & 85.94 & 82.74 & 84.31 & 0.00 \\
HAMMER++~\citep{hammer++}               & 97.60 &  7.99 & 92.26 & 97.72 & 93.70 & 93.34 & 77.66 & 84.12 & 78.62 & 85.86 & 82.89 & 84.35 & -0.09 \\
Qwen2.5VL-72b~\citep{fka-owl}           & 76.67 & 44.93 & 55.06 & \multicolumn{1}{c}{--} & \multicolumn{1}{c}{--} & \multicolumn{1}{c}{--} & \multicolumn{1}{c}{--} & \multicolumn{1}{c}{--} & \multicolumn{1}{c}{--} & \multicolumn{1}{c}{--} & \multicolumn{1}{c}{--} & \multicolumn{1}{c}{--} & \multicolumn{1}{c}{--} \\
FKA-Owl~\citep{fka-owl}                 & 98.09 &  7.19 & 92.60 &  2.53 & 13.97 & 13.84 & 66.40 & 73.54 & 54.82 & 19.16 & 49.71 & 27.66 & -38.01 \\
RamDG~\citep{samm}                      & \underline{98.79} & \underline{5.42} & \underline{94.66} & \underline{98.86} & \underline{95.52} & \underline{95.33} & \underline{80.90} & \underline{87.56} & \underline{82.00} & \underline{86.16} & \underline{83.54} & \underline{84.83} & \underline{+1.80} \\
\midrule
\textbf{REVEAL (Ours)}                  & \textbf{99.83} & \textbf{0.65} & \textbf{97.04} & \textbf{99.56} & \textbf{96.43} & \textbf{96.19} & \textbf{82.69} & \textbf{87.68} & \textbf{82.84} & \textbf{87.25} & \textbf{85.12} & \textbf{86.17} & \textbf{+3.25} \\
\bottomrule
\end{tabular}
}
\end{table*}

\subsection{Objective Function}
Following prior works~\citep{hammer,hammer++}, we define $\mathcal{L}_{\text{Task}}$ as the sum of four task losses:
\begin{equation}
\mathcal{L}_{\text{Task}} =
\mathcal{L}_{\text{BIC}} +
\mathcal{L}_{\text{MLC}} +
\mathcal{L}_{\text{IMG}} +
\mathcal{L}_{\text{TMG}} ,
\end{equation}
where classification objectives are optimized with cross-entropy losses, image grounding with $L_1$ and Generalized IoU losses~\citep{rezatofighi2019generalized}, and text grounding with class-weighted cross-entropy.

To empower the framework with reference-grounded verification and high-recall optimization, we integrate our proposed contrastive, attention-guided, and distillation constraints. The overall optimization objective is formulated as:
\begin{equation}
\mathcal{L}_{\text{total}} = \mathcal{L}_{\text{Task}} + \alpha \mathcal{L}_{\text{VRC}} + \mathcal{L}_{\text{RGAS}} + \beta \mathcal{L}_{\text{distill}},
\end{equation}
where $\mathcal{L}_{\text{VRC}}$ denotes a unified contrastive term consisting of the visual retrieval contrastive loss, $\mathcal{L}_{\text{RGAS}}$ supervises the cross-modal attention patterns for verification, and $\mathcal{L}_{\text{distill}}$ transfers recall-centric knowledge from the teacher expert. $\alpha$ and $\beta$ are trade-off hyperparameters to balance these objectives.

\section{Experiments}

\begin{table*}[t]
\centering
\caption{\textbf{Cross-domain generalization performance (\%)}. By using our comprehensive reference library, models are trained on one news source (indicated by row blocks: BBC, Guardian, USA Today, or Washington Post) and evaluated on all four sources (columns). Gray cells denote in-domain evaluation where training and testing sources match; other cells represent cross-domain transfer scenarios. Best results per column are in \textbf{bold}.
}
\label{tab:cross_domain_optimized}
\setlength{\tabcolsep}{2.5pt}
\resizebox{\textwidth}{!}{
\begin{tabular}{l|c|cccc|cccc|cccc|cccc}
    \toprule
\multirow{2}{*}{\textbf{Methods}} &
\multirow{2}{*}{\diagbox[width=4.5em]{\textbf{Train}}{\textbf{Test}}} &
\multicolumn{4}{c|}{\textbf{BBC}} & \multicolumn{4}{c|}{\textbf{Guardian}} & \multicolumn{4}{c|}{\textbf{USA Today}} & \multicolumn{4}{c}{\textbf{Wash. Post}} \\

\cmidrule(lr){3-6} \cmidrule(lr){7-10} \cmidrule(lr){11-14} \cmidrule(lr){15-18}
& & ACC\textsubscript{cls} & mAP & IoU\textsubscript{m} & ACC\textsubscript{tok} & ACC\textsubscript{cls} & mAP & IoU\textsubscript{m} & ACC\textsubscript{tok} & ACC\textsubscript{cls} & mAP & IoU\textsubscript{m} & ACC\textsubscript{tok} & ACC\textsubscript{cls} & mAP & IoU\textsubscript{m} & ACC\textsubscript{tok} \\
\midrule
HAMMER~\citep{hammer} & \multirow{5}{*}{\rotatebox{90}{\textbf{BBC}}}
& \cellcolor{gray!20}81.75 & \cellcolor{gray!20}79.73 & \cellcolor{gray!20}63.73 & \cellcolor{gray!20}91.47 & 75.76 & 66.13 & 59.69 & 77.44 & 68.11 & 50.84 & 43.94 & 64.91 & 68.21 & 52.69 & 45.60 & 68.48 \\
HAMMER++~\citep{hammer++} &
& \cellcolor{gray!20}81.96 & \cellcolor{gray!20}79.92 & \cellcolor{gray!20}63.77 & \cellcolor{gray!20}90.84 & 75.80 & 66.59 & 60.43 & 77.13 & 67.65 & 50.33 & 43.01 & 64.16 & 68.10 & 52.16 & 45.47 & 68.41 \\
ASAP~\citep{asap} &
& \cellcolor{gray!20}82.48 & \cellcolor{gray!20}80.09 & \cellcolor{gray!20}64.05 & \cellcolor{gray!20}91.91 & 76.69 & 67.71 & 62.89 & 79.63 & 68.94 & 51.58 & 44.81 & 67.58 & 68.51 & 52.50 & 46.52 & 70.46 \\
Our-Top2 &
& \cellcolor{gray!20}85.28 & \cellcolor{gray!20}80.72 & \cellcolor{gray!20}70.30 & \cellcolor{gray!20}92.27 & 83.26 & 70.69 & 69.83 & 84.90 & 79.53 & 59.65 & 63.10 & 80.72 & 81.45 & 59.67 & 65.15 & 82.61 \\
Our-Top1 &
& \cellcolor{gray!20}\textbf{97.09} & \cellcolor{gray!20}\textbf{90.14} & \cellcolor{gray!20}\textbf{83.62} & \cellcolor{gray!20}\textbf{93.90} & \textbf{95.34} & \textbf{82.43} & \textbf{82.90} & \textbf{91.48} & \textbf{93.91} & \textbf{72.42} & \textbf{78.41} & \textbf{90.82} & \textbf{93.66} & \textbf{73.53} & \textbf{78.86} & \textbf{91.63} \\
\midrule
HAMMER~\citep{hammer} & \multirow{5}{*}{\rotatebox{90}{\textbf{Guardian}}}
& 66.65 & 59.80 & 57.30 & 87.22 & \cellcolor{gray!20}87.13 & \cellcolor{gray!20}83.13 & \cellcolor{gray!20}73.53 & \cellcolor{gray!20}89.60 & 66.04 & 59.76 & 36.17 & 87.62 & 66.39 & 59.52 & 35.59 & 88.98 \\
HAMMER++~\citep{hammer++} &
& 65.34 & 59.68 & 56.19 & 87.53 & \cellcolor{gray!20}86.82 & \cellcolor{gray!20}83.29 & \cellcolor{gray!20}73.33 & \cellcolor{gray!20}89.74 & 65.99 & 62.23 & 36.42 & 87.16 & 66.54 & 61.49 & 35.40 & 88.85 \\
ASAP~\citep{asap} &
& 68.65 & 66.23 & 59.22 & 88.84 & \cellcolor{gray!20}88.36 & \cellcolor{gray!20}86.23 & \cellcolor{gray!20}77.86 & \cellcolor{gray!20}91.06 & 69.45 & 65.17 & 40.82 & 87.73 & 69.06 & 65.47 & 41.01 & 89.66 \\
Our-Top2 &
& 81.21 & 64.04 & 70.90 & 89.39 & \cellcolor{gray!20}86.93 & \cellcolor{gray!20}81.26 & \cellcolor{gray!20}72.43 & \cellcolor{gray!20}92.67 & 80.72 & 73.75 & 81.63 & 82.18 & 84.12 & 66.74 & 64.06 & 90.04 \\
Our-Top1 &
& \textbf{95.07} & \textbf{77.78} & \textbf{79.69} & \textbf{91.76} & \cellcolor{gray!20}\textbf{97.70} & \cellcolor{gray!20}\textbf{90.38} & \cellcolor{gray!20}\textbf{83.15} & \cellcolor{gray!20}\textbf{93.41} & \textbf{95.92} & \textbf{85.58} & \textbf{80.72} & \textbf{91.79} & \textbf{95.18} & \textbf{74.4} & \textbf{81.27} & \textbf{91.75} \\
\midrule
HAMMER~\citep{hammer} & \multirow{5}{*}{\rotatebox{90}{\textbf{USA Today}}}
& 70.17 & 52.64 & 45.54 & 82.69 & 74.39 & 59.63 & 50.94 & 83.02 & \cellcolor{gray!20}76.66& \cellcolor{gray!20}70.32 & \cellcolor{gray!20}49.73 & \cellcolor{gray!20}90.9 & 75.77 & 65.33 & 48.38 & 89.95 \\
HAMMER++~\citep{hammer++} &
& 70.25 & 52.36 & 47.07 & 82.29 & 74.84 & 59.17 & 52.07 & 82.22 & \cellcolor{gray!20}76.80 & \cellcolor{gray!20}71.27 & \cellcolor{gray!20}50.72 & \cellcolor{gray!20}91.15 & 75.85 & 66.30 & 50.30 & 90.11 \\
ASAP~\citep{asap} &
& 74.02 & 56.25 & 50.95 & 81.44 & 79.91 & 63.58 & 57.68 & 82.82 & \cellcolor{gray!20}86.10 & \cellcolor{gray!20}75.63 & \cellcolor{gray!20}56.02 & \cellcolor{gray!20}92.64 & 83.95 & 70.67 & 54.97 & 91.00 \\
Our-Top2 &
& 80.96 & 63.24 & 65.15 & 87.30 & 82.22 & 66.55 & 69.20 & 86.80 & \cellcolor{gray!20}84.42 & \cellcolor{gray!20}79.14 & \cellcolor{gray!20}68.14 & \cellcolor{gray!20}92.16 & 85.32 & 75.10 & 67.43 & 92.13 \\
Our-Top1 &
& \textbf{94.35} & \textbf{75.62} & \textbf{79.29} & \textbf{91.35} & \textbf{94.97} & \textbf{78.46} & \textbf{83.84} & \textbf{91.31} & \cellcolor{gray!20}\textbf{97.25} & \cellcolor{gray!20}\textbf{87.41} & \cellcolor{gray!20}\textbf{82.5} & \cellcolor{gray!20}\textbf{93.13} & \textbf{96.93} & \textbf{84.17} & \textbf{81.59} & \textbf{93.59} \\
\midrule
HAMMER~\citep{hammer} & \multirow{5}{*}{\rotatebox{90}{\textbf{Wash. Post}}}
& 70.58 & 53.72 & 44.61 & 83.10 & 74.10 & 58.8 & 50.27 & 83.36 & 74.80 & 66.84 & 47.22 & 89.71 & \cellcolor{gray!20}75.74 & \cellcolor{gray!20}65.71 & \cellcolor{gray!20}48.99 & \cellcolor{gray!20}89.84 \\
HAMMER++~\citep{hammer++} &
& 70.93 & 53.02 & 44.66 & 82.23 & 75.03 & 59.82 & 50.52 & 83.50 & 75.49 & 68.08 & 47.45 & 89.01 & \cellcolor{gray!20}75.85 & \cellcolor{gray!20}67.50 & \cellcolor{gray!20}48.52 & \cellcolor{gray!20}89.89 \\
ASAP~\citep{asap} &
& 71.86 & 57.28 & 49.75 & 82.77 & 80.71 & 63.60 & 56.29 & 84.04 & 77.58 & 71.59 & 53.00 & 89.25 & \cellcolor{gray!20}78.57 & \cellcolor{gray!20}73.10 & \cellcolor{gray!20}54.10 & \cellcolor{gray!20}91.91 \\
Our-Top2 &
& 82.24 & 61.16 & 61.70 & 87.06 & 82.54 & 65.34 & 65.28 & 86.43 & 80.85 & 72.94 & 61.07 & 90.88 & \cellcolor{gray!20}81.73 & \cellcolor{gray!20}75.37 & \cellcolor{gray!20}63.35 & \cellcolor{gray!20}91.33 \\
Our-Top1 &
& \textbf{95.89} & \textbf{76.46} & \textbf{78.60} & \textbf{91.55} & \textbf{95.63} & \textbf{79.05} & \textbf{81.70} & \textbf{90.73} & \textbf{96.98} & \textbf{84.16} & \textbf{80.68} & \textbf{91.38} & \cellcolor{gray!20}\textbf{97.53} & \cellcolor{gray!20}\textbf{86.37} & \cellcolor{gray!20}\textbf{79.58} & \cellcolor{gray!20}\textbf{92.88} \\
\bottomrule
\end{tabular}
}
\end{table*}

\begin{table*}[t]
\centering
\caption{\textbf{Cross-dataset generalization on MDSM (\%)}. Models are trained on the Guardian domain of DGM$^4$ and evaluated using our comprehensive reference library. \textbf{Bold} indicates best and \underline{underlined} indicates second-best.}
\label{tab:cross_dataset_mdsm}
\setlength{\tabcolsep}{4pt}
\renewcommand{\arraystretch}{1.15}
\small
\resizebox{\textwidth}{!}{
\begin{tabular}{@{}lccccccccccccccccccc@{}}
\toprule
\multirow{2}{*}{\textbf{Methods}} &
\multicolumn{3}{c}{\textbf{Guardian}} &
\multicolumn{3}{c}{\textbf{BBC}} &
\multicolumn{3}{c}{\textbf{NYT}} &
\multicolumn{3}{c}{\textbf{USA Today}} &
\multicolumn{3}{c}{\textbf{Wash.\ Post}} &
\multicolumn{3}{c}{\textbf{AVG}} \\
\cmidrule(lr){2-4} \cmidrule(lr){5-7} \cmidrule(lr){8-10} \cmidrule(lr){11-13} \cmidrule(lr){14-16} \cmidrule(lr){17-19}
& ACC & mAP & mIoU
& ACC & mAP & mIoU
& ACC & mAP & mIoU
& ACC & mAP & mIoU
& ACC & mAP & mIoU
& ACC & mAP & mIoU \\
\midrule
HAMMER~\citep{hammer}       & 73.51 & 44.06 & 60.78 & 51.53 & 38.44 & 63.21 & 65.33 & 25.23 & 38.94 & 59.17 & 31.78 & 28.14 & 67.81 & 33.89 & 27.48 & 63.47 & 34.68 & 43.71 \\
HAMMER++~\citep{hammer++}   & 73.57 & \underline{45.86} & 60.74 & 50.72 & 39.12 & \underline{63.97} & 65.21 & 25.09 & 38.22 & 59.84 & 33.73 & 28.71 & 67.62 & 35.30 & 28.40 & 63.39 & 35.82 & 44.01 \\
FKA-Owl~\citep{fka-owl}     & 69.18 & 19.46 & 15.01 & \underline{57.94} & 22.94 & 11.37 & 61.29 & 18.03 & 12.27 & 58.95 & 30.68 & 16.74 & 62.30 & 35.15 & 18.01 & 61.93 & 25.25 & 14.68 \\
RamDG~\citep{samm}          & 72.49 & 43.84 & 60.69 & 55.27 & \underline{41.25} & 63.50 & 65.55 & 25.78 & 40.28 & 66.82 & 34.33 & 28.23 & 66.15 & 36.17 & 27.72 & 65.26 & \underline{36.27} & 44.08 \\
ASAP~\citep{asap}           & \underline{73.63} & 44.91 & 61.25 & 51.55 & 38.97 & 63.74 & \underline{66.13} & 26.29 & 39.67 & 59.89 & 33.47 & 29.87 & 67.71 & 35.44 & 29.61 & 63.78 & 35.82 & 44.83 \\
AMD~\citep{amd}             & 69.89 & 33.67 & \textbf{73.17} & 52.68 & 18.39 & 62.11 & 64.62 & \underline{29.89} & \textbf{41.57} & \textbf{73.28} & \underline{37.04} & \underline{50.89} & \textbf{73.24} & \underline{39.61} & \underline{50.94} & \underline{66.74} & 31.72 & \underline{55.74} \\
\midrule
\textbf{REVEAL (Ours)}      & \textbf{78.52} & \textbf{50.28} & \underline{65.31} & \textbf{72.69} & \textbf{55.22} & \textbf{76.66} & \textbf{70.22} & \textbf{31.56} & \underline{40.56} & \underline{66.62} & \textbf{51.46} & \textbf{63.95} & \underline{70.51} & \textbf{51.92} & \textbf{67.08} & \textbf{71.71} & \textbf{48.09} & \textbf{62.71} \\
\bottomrule
\end{tabular}
}
\end{table*}

\paragraph{Experimental Setup.}
For reference-grounded verification, we build an inference-time library from VisualNews~\citep{visualnews}, containing 170K authentic image--text pairs covering over 40K public figures. 

\subsection{Quantitative Results}
\label{sec:quantitative_results}
We evaluate REVEAL from four perspectives: in-domain performance on DGM$^4$~\citep{hammer}, robustness on SAMM~\citep{samm}, cross-domain generalization across news sources, and zero-shot cross-dataset transfer to MDSM.

\noindent\textbf{(1) Superiority on DGM$^4$ (Tab.~\ref{tab:DGM}).}
By using our comprehensive reference library, REVEAL achieves consistent improvements across all four task families, including binary classification, multi-label classification, image grounding, and text grounding. 
It obtains an average gain of +6.92\% over HAMMER~\citep{hammer}, with 97.82\% AUC and 93.18\% ACC for binary classification. 
Notably, REVEAL reduces EER from 14.10\% to 3.75\% compared with HAMMER, showing that reference-grounded comparison provides more reliable authenticity judgments than artifact-only detection. 
Compared with IDseq~\citep{idseq}, REVEAL achieves stronger overall performance, and it substantially outperforms ASAP~\citep{asap} on binary classification and image grounding, demonstrating the advantage of retrieval-augmented verification beyond cross-modal alignment.

\noindent\textbf{(2) Robustness on SAMM (Tab.~\ref{tab:comparison}).}
To further evaluate robustness beyond DGM$^4$, we benchmark REVEAL on SAMM~\citep{samm}, which is constructed with different manipulation pipelines. 
REVEAL establishes a new state of the art across all metrics, achieving 99.83\% AUC, 0.65\% EER, and 97.04\% ACC. 
Compared with the strongest baseline RamDG~\citep{samm}, REVEAL reduces EER from 5.42\% to 0.65\% and improves IoU\textsubscript{m} by 1.79 points, indicating that authentic reference anchors remain effective under different manipulation distributions.

\noindent\textbf{(3) Cross-Domain Generalization (Tab.~\ref{tab:cross_domain_optimized}).}
To evaluate domain robustness, we train models on a single news source and evaluate them across all four sources. To simulate a challenging, suboptimal retrieval scenario, we introduce a variant (Our-Top2) that intentionally excludes the top-1 retrieved reference. Despite this restriction, it consistently outperforms HAMMER~\citep{hammer}, HAMMER++~\citep{hammer++}, and ASAP~\citep{asap}, achieving average improvements of +10.2\% in ACC and +15.8\% in $\text{IoU}_m$ over ASAP. Furthermore, when the full gallery is accessible (Our-Top1), performance improves even further. Crucially, this gain is achieved simply by updating the retrieval index without any parameter tuning, confirming that our plug-and-play memory effectively enables training-free domain adaptation.

\noindent\textbf{(4) Zero-Shot Cross-Dataset Transfer (Tab.~\ref{tab:cross_dataset_mdsm}).}
We further test generalization under a harder setting by training models only on the Guardian split of DGM$^4$ and evaluating them on MDSM without fine-tuning. 
REVEAL achieves the best average results with 71.71\% ACC, 48.09\% mAP, and 62.71\% mIoU, surpassing the next-best baseline by +4.97\%, +11.82\%, and +6.97\%, respectively. 
This demonstrates that reference-grounded verification generalizes across different news sources, manipulation pipelines, and annotation protocols.

\begin{table*}[t]
\centering
\caption{Ablation study of different components on the DGM$^4$ dataset using our comprehensive reference library. $I_{\mathrm{ref}}$: Visual Reference Stream; $T_{\mathrm{ref}}$: Text Reference Stream; $T_{\text{moe}}$: Reference-Distilled Task-Adaptive Experts. Baseline denotes the model without any retrieval augmentation.}
\label{tab:overall_ablation}
\setlength{\tabcolsep}{4pt}
\renewcommand{\arraystretch}{1.1}
\small
\resizebox{\textwidth}{!}{
\begin{tabular}{@{}l S[table-format=2.2] S[table-format=2.2] S[table-format=2.2] S[table-format=2.2] S[table-format=2.2] S[table-format=2.2] S[table-format=2.2] S[table-format=2.2] S[table-format=2.2] S[table-format=2.2] S[table-format=2.2] S[table-format=2.2]@{}}
\toprule
\multirow{2}{*}{\textbf{Variants}} &
\multicolumn{3}{c}{\textbf{Binary Cls}} &
\multicolumn{3}{c}{\textbf{Multi-Label Cls}} &
\multicolumn{3}{c}{\textbf{Image Grounding}} &
\multicolumn{3}{c}{\textbf{Text Grounding}} \\
\cmidrule(lr){2-4} \cmidrule(lr){5-7} \cmidrule(lr){8-10} \cmidrule(lr){11-13}
& {AUC $\uparrow$} & {EER $\downarrow$} & {ACC $\uparrow$}
& {mAP $\uparrow$} & {CF1 $\uparrow$} & {OF1 $\uparrow$}
& {IoU\textsubscript{m} $\uparrow$} & {IoU\textsubscript{50} $\uparrow$} & {IoU\textsubscript{75} $\uparrow$}
& {Precision $\uparrow$} & {Recall $\uparrow$} & {F1 $\uparrow$} \\
\midrule
Baseline  & 90.07 & 15.84 & 84.78 & 83.23 & 78.69 & 78.84 & 79.21 & 86.93 & 74.72 & 71.70 & 66.02 & 68.74 \\
w/ $I_{\mathrm{ref}}$  & 95.78 & 5.15 & 91.75 & 89.59 & 82.44 & 83.22 & 84.14 & 92.30 & 80.82 & 74.96 & 64.13 & 69.12 \\
w/ $I_{\mathrm{ref}}$+$T_{\mathrm{ref}}$  & 96.66 & 4.31 & 92.91 & 89.95 & 83.54 & 84.14 & 84.21 & 92.39 & 81.40 & \textbf{82.84} & 68.83 & 75.15 \\
w/ $I_{\mathrm{ref}}$+$T_{\mathrm{ref}}$+$T_{\text{moe}}$ & \textbf{97.82} & \textbf{3.75} & \textbf{93.18} & \textbf{91.37} & \textbf{85.13} & \textbf{86.68} & \textbf{85.51} & \textbf{93.53} & \textbf{86.52} & 79.31 & \textbf{74.45} & \textbf{75.34} \\
\bottomrule
\end{tabular}
}
\end{table*}

\subsection{Ablation Study}
We conduct comprehensive ablation experiments to dissect the contribution of each component in REVEAL.
The baseline is a standard BEiT-3 model that relies solely on internal artifact memorization.
To validate our reference-grounded paradigm, we systematically integrate the two parallel branches of our Authenticity Conditioned Cross Attention (ACCA) module.
We first activate the Visual Reference Stream, which processes retrieved $I_{\mathrm{ref}}$ via the visual discrepancy modeling branch.
Subsequently, we enable the Text Reference Stream, which leverages the textual consistency verification branch to align query texts with retrieved $T_{\mathrm{ref}}$.
Finally, we orchestrate these reference-enhanced features via the Task-Aware MoE.
Table~\ref{tab:overall_ablation} summarizes the progressive improvements.

\paragraph{Visual Reference Stream ($I_{\mathrm{ref}}$).}
Activating the visual reference stream introduces authentic image anchors to the visual discrepancy modeling branch of ACCA.
This integration yields substantial gains across all metrics, with AUC improving from 90.07\% to 95.78\% (+5.71\%) and Image Grounding (IoU\textsubscript{m}) increasing from 79.21\% to 84.14\% (+4.93\%).
These results validate that the ACCA module effectively transforms the detection paradigm. By computing feature-level residuals against the retrieved $I_{\mathrm{ref}}$, the model shifts from searching for absolute artifacts to measuring relative inconsistencies, thereby significantly enhancing localization precision.

\paragraph{Text Reference Stream ($T_{\mathrm{ref}}$).}
Complementing the visual pathway, the text reference stream integrates authentic captions through the textual consistency verification branch of ACCA.
This transforms the model into a complete dual-stream framework, resulting in a pronounced improvement in text-related metrics (Text Grounding F1: +6.03\%).
The performance surge is attributed to the semantic anchoring mechanism within ACCA. By employing cross-attention to contrast query texts against verified logic in $T_{\mathrm{ref}}$, the model becomes capable of exposing subtle semantic contradictions that purely statistical language models fail to detect.

\begin{figure}[t]
\centering
\begin{subfigure}[t]{0.48\columnwidth}
    \centering
    \includegraphics[width=\linewidth]{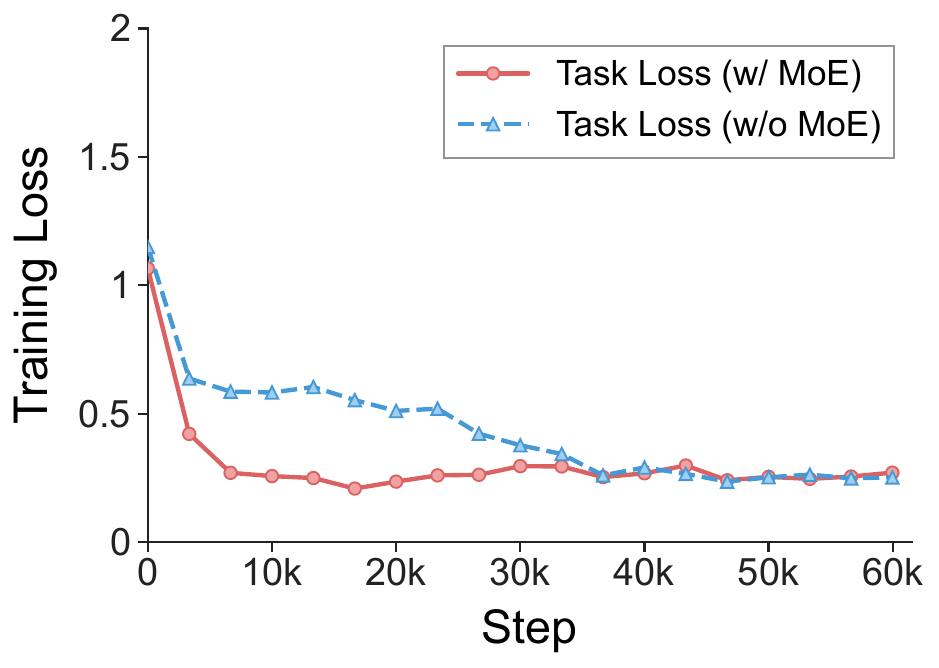}
    \caption{Task Loss}
    \label{fig:task_loss}
\end{subfigure}
\hfill
\begin{subfigure}[t]{0.48\columnwidth}
    \centering
    \includegraphics[width=\linewidth]{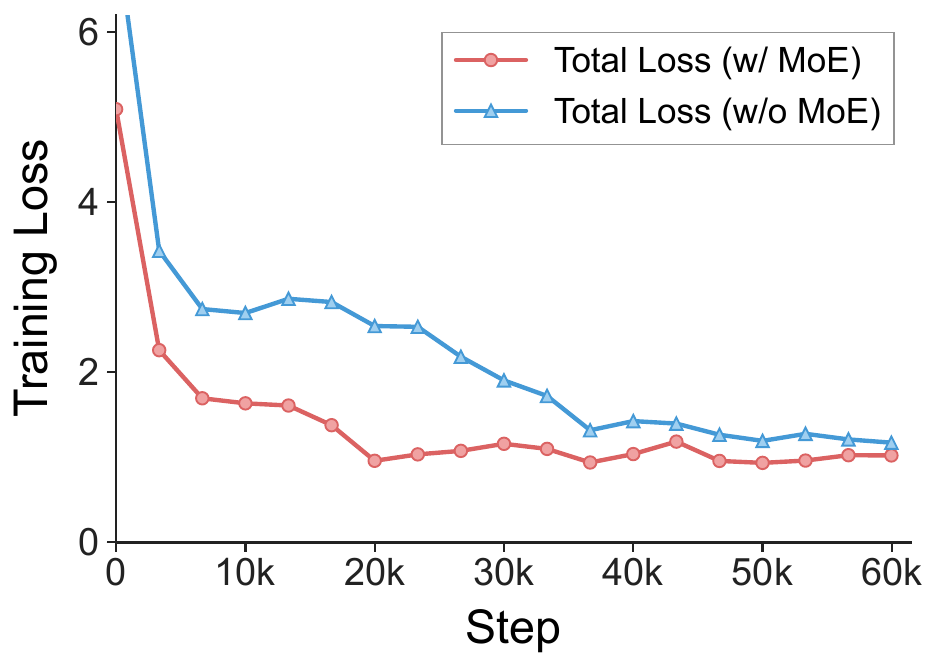}
    \caption{Total Loss}
    \label{fig:total_loss}
\end{subfigure}
\caption{Training dynamics on the DGM$^4$ dataset. The curves (Task Loss and Total Loss) demonstrate that incorporating the proposed MoE detection head leads to faster convergence compared to the baseline.}
\label{fig:task_total_loss_compare}
\end{figure}

\paragraph{Impact of Task-Aware MoE ($T_{\text{moe}}$).}
The MoE architecture with teacher expert distillation provides additional performance gains across all metrics. As discussed in Section~\ref{sec:method}, global classification and local grounding tasks impose different optimization pressures on shared representations. The MoE design addresses this by routing features to specialized experts, allowing each task to leverage tailored feature transformations while still benefiting from shared low-level representations. The consistent improvements across both classification and grounding metrics confirm that this architecture successfully mitigates task interference.

Figure~\ref{fig:task_total_loss_compare} compares the training dynamics of REVEAL with and without the task-decoupling MoE detection head. The MoE-equipped model shows faster optimization in the early training stage and consistently reaches a lower total loss throughout training. For the task loss, both variants eventually converge to comparable values, while the MoE variant descends more rapidly and remains more stable after convergence. These trends suggest that task-adaptive expert routing facilitates optimization under heterogeneous objectives, allowing global classification and fine-grained grounding tasks to share representations more effectively while reducing optimization interference.

\section{Conclusion}
We present REVEAL, a retrieval augmented framework that reformulates multimodal manipulation detection as reference-grounded verification.
By retrieving authentic references and modeling query-reference discrepancies, REVEAL shifts from isolated artifact memorization to comparative reasoning.
Extensive experiments demonstrate state-of-the-art performance across detection and grounding tasks.
Notably, our paradigm enables training-free domain adaptation by simply updating the reference library, achieving strong cross-domain transferability without parameter updates.
The reference-grounded paradigm offers a complementary perspective to artifact-based detection, more closely aligning with human fact-checking processes while providing interpretable evidence for detection decisions.

\section*{Ethics Statement}
All datasets utilized in this study are derived from publicly available sources and legitimate news media. We strictly ensure that the collected data exclusively features public figures in public spheres, thereby avoiding any privacy infringement of private individuals. Furthermore, we explicitly declare that all data and the proposed models are restricted to non-commercial, scientific research purposes aimed at combating multimodal misinformation.

\section*{Limitations}
Like other retrieval-augmented methods, its performance hinges on the coverage and quality of the reference library. For rare or out-of-distribution entities, retrieved neighbors may be uninformative, degrading the reference-grounded paradigm under retrieval failures. 
Although the reference gallery can be constructed offline, our implementation builds the gallery at a speed of 697 images per second; when the gallery is large or needs to be frequently updated, this additional preprocessing step still introduces certain computational overhead.
Moreover, due to space limitations in the main text, we moved several supplementary experiments and analyses to the appendix, which may slightly affect the overall reading experience.

\bibliography{cite}
\clearpage
\appendix
\raggedbottom

\end{document}